\documentclass[runningheads,orivec]{llncs}
\usepackage[T1]{fontenc}
\usepackage{graphicx}
\usepackage[misc,geometry]{ifsym}

\usepackage{amsmath}
\usepackage{textcomp}
\usepackage{gensymb}
\usepackage[table,xcdraw]{xcolor}
\usepackage{svg}
\graphicspath{ {./Images/} }
\usepackage{hyperref}
\usepackage{orcidlink}

\usepackage{algorithm}
\usepackage{algpseudocode}
\usepackage{makecell}

\usepackage{siunitx}
\begin{document}
\title{Robot Localization Using a Learned Keypoint Detector and Descriptor with a Floor Camera and a Feature Rich Industrial Floor}
\titlerunning{Robot Localization with Learned Keypoint Detection on Industrial Floors}
%
\author{Piet Brömmel\inst{2}, Dominik Brämer\textsuperscript{(\Letter)}\orcidlink{0009-0003-9326-432X}\inst{1}, Oliver Urbann\orcidlink{0000-0001-8596-9133}\inst{2} and Diana Kleingarn\orcidlink{0009-0001-1751-0504}\inst{1}}
\authorrunning{P. Brömmel, D. Brämer,  O. Urbann \and D. Kleingarn}
%
\institute{Robotics Research Institute, Section Information Technology,\\
TU Dortmund University, 44227 Dortmund, Germany \and
Fraunhofer IML,\\ Joseph-von-Fraunhofer-Str. 2-4, Dortmund, Germany\\
\email{dominik.braemer@tu-dortmund.de}}
\maketitle              

\begin{abstract}

The localization of moving robots depends on the availability of good features from the environment. 
Sensor systems like Lidar are popular, but unique features can also be extracted from images of the ground.
This work presents the Keypoint Localization Framework (KOALA), which utilizes deep neural networks that extract sufficient features from an industrial floor for accurate localization without having readable markers. 
For this purpose, we use a floor covering that can be produced as cheaply as common industrial floors.
Although we do not use any filtering, prior, or temporal information, we can estimate our position in \SI{75.7}{\percent} of all images with a mean position error of \SI{2}{\centi\metre} and a rotation error of \SI{2.4}{\percent}.
Thus, the robot kidnapping problem can be solved with high precision in every frame, even while the robot is moving.
Furthermore, we show that our framework with our detector and descriptor combination is able to outperform comparable approaches.
\end{abstract}
\section{Introduction}
\label{sec:intro}

Indoor localization is crucial for autonomous mobile robots, making it a key research topic and of significant commercial interest.
The current position is essential for various applications, leading to multiple approaches.
GPS is a precise and widely known solution that is unsuitable for indoor applications. 
Ultra-wideband requires expensive hardware to be mounted in the whole location. 
As a result, these methods are rarely used in mobile robots.

A popular approach is using light detection and ranging (Lidar) sensors together with a simultaneous localization and mapping (SLAM) method. 
The price range for these sensors vary widely and with it the range of capabilities, e.g., the maximum measurable distance and speed of rotation.

Another popular class of methods is the feature detection of the environment with cameras.
These can be QR codes stuck on the floor, ceiling, or other artificial features placed in the environment. 
It is also possible to detect features that are naturally present. 

However, sensing static features in dynamic environments with many moving objects (e.g., other robots or people) is generally challenging.
Therefore, a typical viewing direction for cameras is towards the ceiling~\cite{zhang2015}.
Occlusions and the large distance to the roof can be tricky and make it hard to detect smaller features.

In this paper, we investigate the possibility of mounting the camera beneath the robot, allowing for a close-up view of the ground.
This eliminates occlusions, particularly from moving objects, and ensures uniform illumination.
For this reason, we present a floor image localization framework based on a custom keypoint detector and descriptor.

\begin{figure}[t]
    \centering
    \includegraphics[width=0.6\textwidth]{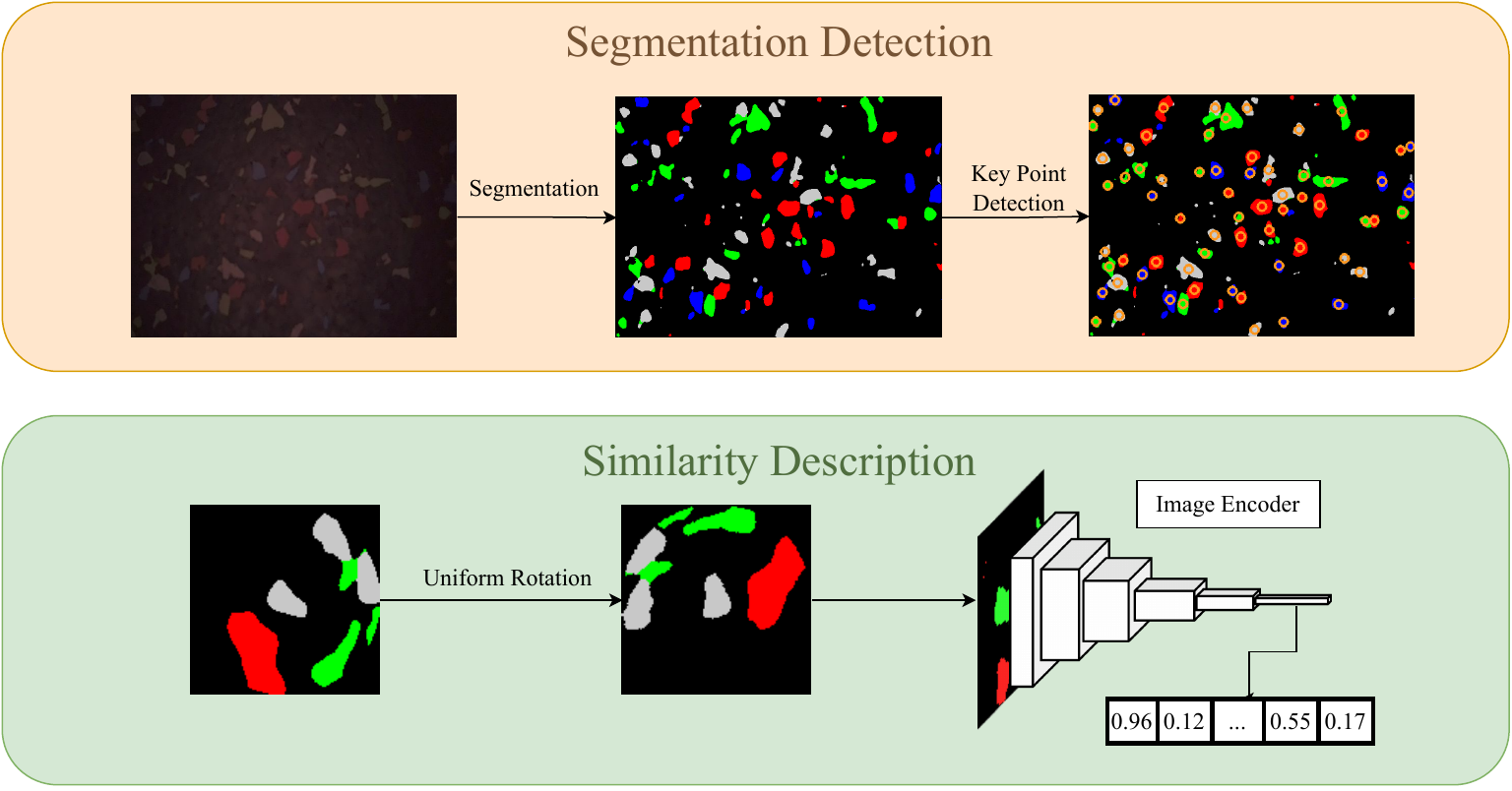}
    \caption{
    A conceptual overview of our segmentation detector and similarity descriptor. Upper: Raw image segmented into an RGBW mask with detected keypoints. Lower: Keypoint patches are rotated uniformly and encoded into latent vectors by a pretrained encoder.
    }
    \label{fig:pipeline_overview}
    \vspace{-15px}
\end{figure}
Possible features to be detected are QR codes for driving predefined paths~\cite{teja2018}. 
A dense mesh of QR codes is needed to use them on the ground to localize beyond predefined paths.

\begin{figure*}[t]
    \centering
    \includegraphics[width=0.3\textwidth]{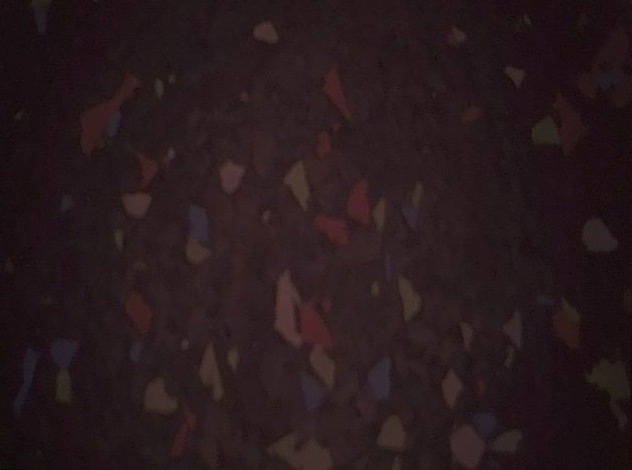} 
    \includegraphics[width=0.3\textwidth]{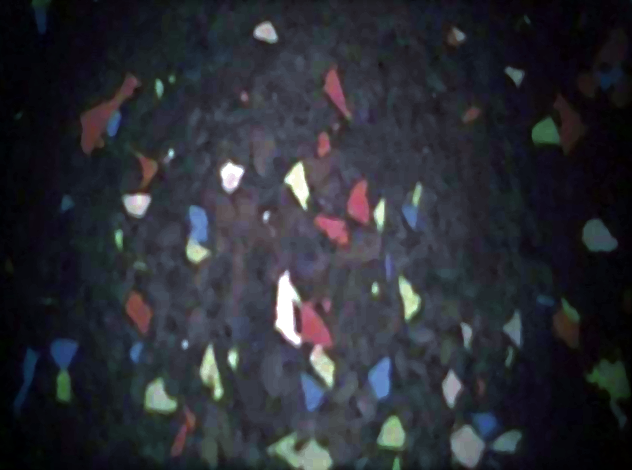} 
    \includegraphics[width=0.3\textwidth]{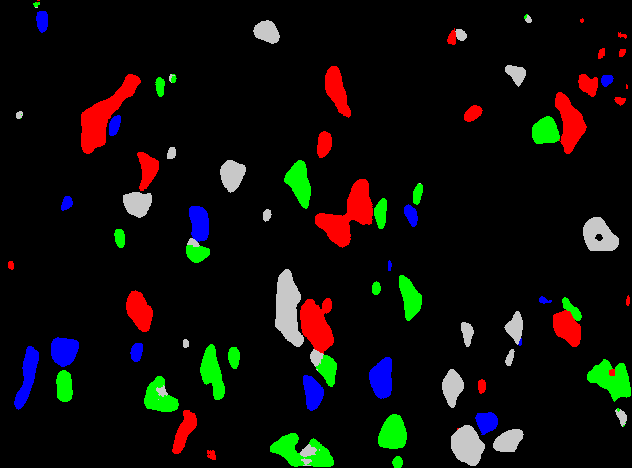} 
    \caption{
    Left, a raw \qtyproduct{4.95 x 2.80}{\centi\metre} floor image with RGBW pattern.
    The same image brightened and contrasted for improved clarity in the middle, and the segmentation mask of the image on the right.
    }
    \label{fig:ground}
    \vspace{-15px}
\end{figure*}

At the other extreme are popular floor coverings such as concrete or asphalt~\cite{chen2018streetmap}. 
In this work, we choose a trade-off between these two poles and use industrial flooring typical for robotics. 
In its production, black granulate is used as a base color mixed with red, green, blue, and white colored granulates.
This creates a random red, green, blue, and white (RGBW) pattern on a black background that is clearly visible, see \autoref{fig:ground}. 
We use this pattern for localization with a preceding phase for mapping.

\section{Related Work}
\label{related_work}

As mentioned above, methods for simultaneous localization and mapping (SLAM) based on cameras or laser scanners are very popular~\cite{moosmann2011velodyne,mur2017orb}. 
Other examples of localization methods use cameras pointed at the ceiling as proposed by Hwang et al. and Chen et al.~\cite{chen2014indoor,hwang2013monocular}, who also confirm the disadvantages mentioned in \autoref{sec:intro}.

A few publications already address the application of ground cameras.
In general, it can be shown that methods like SIFT and \mbox{CenSure} are well suited to extract features from patterns on the ground~\cite{schmid2020features}.

Kozak et al. propose a method with commercially available hardware to perform map-based localization using a ground-facing camera~\cite{kozak2016ranger}.
For their best localization results, they use a combination of \mbox{CenSure} and SIFT as feature detector and descriptor, but they operate in productive use a combination of \mbox{CenSure} and ORB for a faster execution time.
They apply it to asphalt and report a good lateral accuracy within \SI{2}{\centi\metre}, but without exact numbers due to missing ground truth values. 
The system proposed by Kozak et al. depends on previous positions to deliver a correct localization through an extended Kalman filter.
Therefore, they need an initial starting position and cannot solve the robot kidnapping problem.

Similarly, Zhang et al. and Schmid et al. have developed a high-precision localization that works on different floor types, such as asphalt, concrete, tiles, and carpet~\cite{SchmidSM20,zhang2019high}.
They use SIFT as a feature detector and SIFT or LATCH as a descriptor for their best localization results.
Instead of position prediction, they perform image retrieval for evaluation, due to missing ground truth data.

In particular, CNN encoders, as an approach from the field of machine learning, are a popular method to learn a compressed representation for a set of data~\cite{SIFT_Meets_CNN}.
This representation can be utilized to find similar images~\cite{en2017unsupervised}.

SIFT is particularly successful because of Lowe's matching criterion. In order to make this advantage available for deep neural networks, Mishchuk et al. propose a loss that maximizes the distance between the closest positive and closest negative example~\cite{mishchuk2018working}.

\begin{figure}[ht]
    \centering
    \medskip    
    \vspace{-15px}
    \begin{minipage}[b]{0.5\textwidth}
    \centering
    \includegraphics[width=0.9\textwidth]{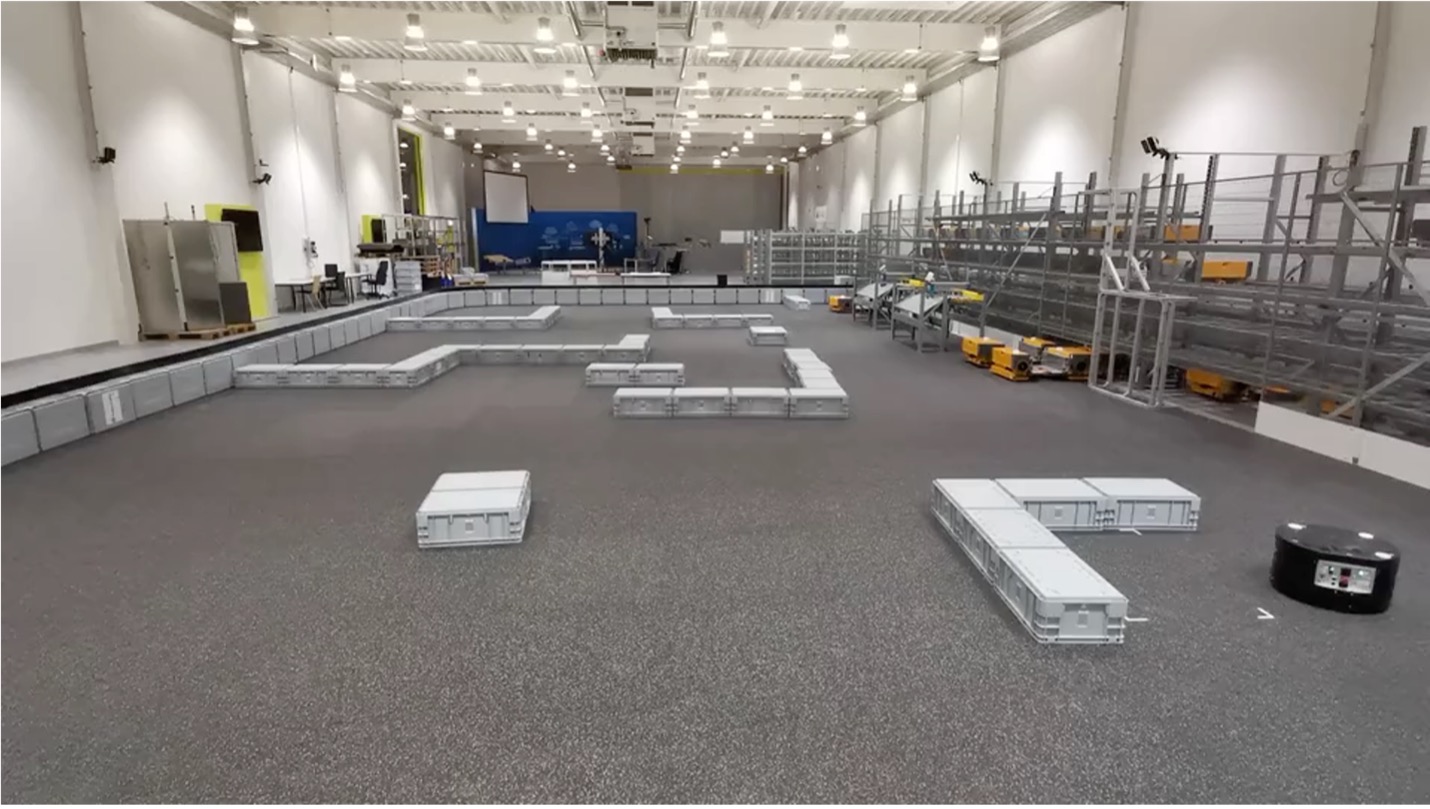}
    \caption{Overview of the experimentation hall showing the motion capture system and the industrial floor.}
    \label{fig:hall}
    \end{minipage}
    \hfill
    \begin{minipage}[b]{0.45\textwidth}
    \centering
    \includegraphics[width=0.85\textwidth]{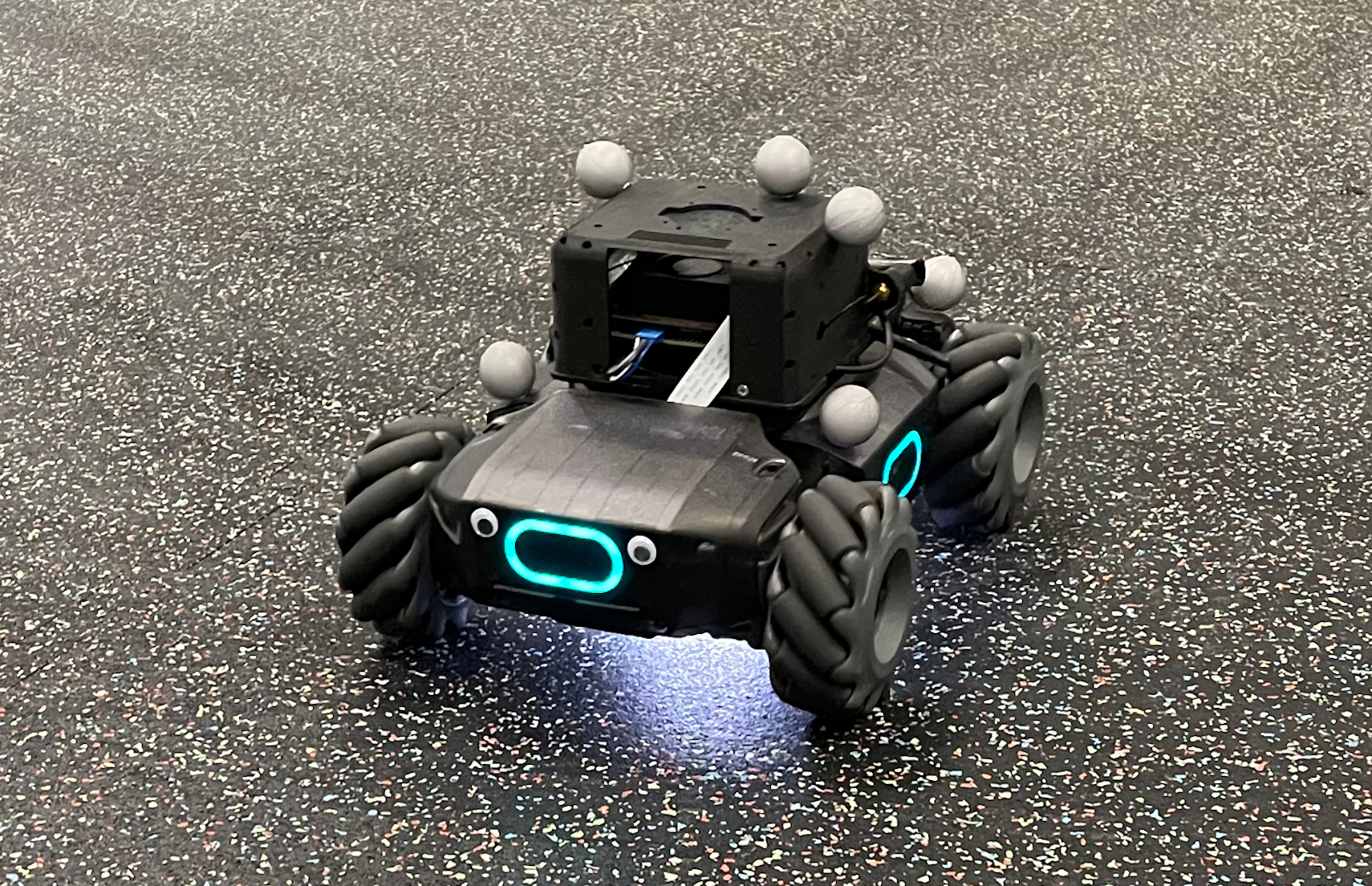} 
    \caption{Modified DJI Robomaster S1 with a floor camera and markers for the motion capture system.}
    \label{fig:robot}
    \end{minipage}
    \vspace{-15px}
\end{figure}

Zhang et al. utilize this to apply an autoencoder that generates descriptions of selected parts (keypoints) of an image. 
They apply this method to reidentify images of textured floors~\cite{zhang2018learning}. 
By combining this with a SIFT descriptor, they motivate the application as a localization method. 
However, this application is only briefly introduced, and the localization error against a ground truth position is not evaluated. 

In contrast, Chen et al.~\cite{chen2018streetmap} present a complete localization pipeline and report localization errors on various floors: a few millimeters on floors with visible lines (e.g., tiled floors) and \SI{10}{\centi\metre} up to \SI{13}{\centi\metre} on general floors. 
However, as lines cannot be identified uniquely and sometimes no line is visible, the solution for tiled floors is combined with an extended Kalman filter, and in some cases, a different system has to provide an initial localization. 
Furthermore, rotation localization error is not evaluated as no ground truth data exists.

\subsection{Contribution}
As can be seen in the related research, localization on general soils still poses a challenge. Therefore, this paper proposes a localization framework that
\begin{itemize}
    \item uses an industrial floor made of random colored plastic granulate for position prediction,
    \item uses a custom keypoint detector and descriptor,
    \item provides a localization with a mean distance error of around \SI{2}{\centi\metre},
    \item does not use temporal information,
    \item is evaluated utilizing a motion capture system.
\end{itemize}

Furthermore, we use our dataset with ground truth positions consisting of 1.2 million images (\SI{280}{\giga\byte}) described in \autoref{dataset} to train, test and evaluate our approach.
We also use it to benchmark our approach against other comparable approaches of Kozak et al.~\cite{kozak2016ranger} and Zhang et al.~\cite{zhang2019high}  that have not been studied with this accuracy before.

\subsection{Structure}

We begin by introducing the dataset of overlapping floor images in \autoref{dataset} which covers an area of \SI{144}{\metre\squared}, and the evaluation runs.
In \autoref{sec:sysdes}, we present our approach for the detector and descriptor, along with our localization framework (KOALA). 
This framework includes two main components: map creation and position estimation, both of which are explained in \autoref{sec:sysdes}.
Finally, in \autoref{sec:eval}, we evaluate the performance of KOALA using our detector and descriptor compared to other feature extraction methods discussed in \autoref{related_work}.

\section{Dataset}
\label{dataset}
The dataset for this work was recorded in our lab due to a lack of available datasets.
A thorough search showed that only Zhang et al. published a dataset with different floor types. 
However, the ground truth position information were not part of this dataset~\cite{zhang2019high}.
To provide ground-truth position information for each image in our dataset, we use a motion capture system. 
\autoref{fig:hall} shows an overview of the hall, the motion capture system, and the area covered with the industrial floor.

We use cameras from the company Vicon, model V5, V8, and V16, for which the robot is equipped with markers, see \autoref{fig:robot}. 
The images of the ground are captured at a frequency of \SI{60}{\hertz} and the position of the robot on the field at a frequency of \SI{200}{\hertz}.
We synchronize the images and positions to gain our training and evaluation dataset with a precision of \SI{0.5}{\centi\metre}.

The images are recorded using a Raspberry Pi Camera Module 2 and an LED ring.
Lens distortion correction is applied to every image. 
The camera captures an area of \qtyproduct{4.95 x 2.80}{\centi\metre} with \qtyproduct{632 x 480}{px}.
An example image of the floor can be seen in \autoref{fig:ground}.

The dataset consists of 36 mapping runs of \qtyproduct{2 x 2}{\metre} covering \SI{144}{\metre\squared} and 12 evaluation runs.
The evaluation runs are split into 4 groups covering \SI{4}{\metre\squared}, \SI{36}{\metre\squared} and \SI{144}{\metre\squared}.
The mapping runs are captured with a robot speed of \SI{0.2}{\metre/\second} for more overlapping mapping images. 
In contrast, the evaluation runs are captured with \SI{0.3}{\metre/\second}, which was the maximum travel speed without a high degree of motion blur.

\section{System Design}
\label{sec:sysdes}

This section presents a custom Segmentation Detector (SEG) and Similarity Descriptor (SIM).
Furthermore, we propose our localization framework consisting of a map creation and position estimation step.

\subsection{Detector}
Relevant features for this floor are red, green, blue, and white color blobs, which can be obscured by sensor noise and lighting conditions (\autoref{fig:ground}). 
To filter this noise, our keypoint detector first employs a U-Net Xception-style segmentation model that learns a mask for the four colors, as this architecture could achieve good results with a low runtime~\cite{chollet2017xception,ronneberger2015unet}.
Given the local nature of these features, we reduce the network to two stages with an 8-filter size to boost speed and generalization. 
We manually annotated 20 images from our dataset (16 for training, 4 for validation) and used flip, crop, shear, translate, and rotate augmentations to extend our training and validation images.
Training for 4000 iterations with a batch size of 64 yielded a final sparse categorical cross-entropy validation loss of 0.072. 

In the second step, color blobs are extracted from the segmented image using a connected component labeling algorithm based on Fiorio et al.~\cite{fiorio1996two,wu2005optimizing}.
The center of a color blob is used as a keypoint if
\begin{enumerate}
    \item the center is \SI{64}{px} away from the image border,
    \item the area of the color blobs contains more than \SI{150}{px} and
    \item there are at least 500 colored pixel in a radius of \SI{64}{px} around the center.
\end{enumerate}

\subsection{Descriptor}
We extract a rotation-invariant descriptor from a circular region with a radius of \SI{64}{px} around each keypoint by first rotating the image and then encoding it with a CNN. 
To achieve uniform rotation, we approximate each color blob with an ellipse and align the patch according to its major axis, ensuring the upper half contains more colored pixels—a necessary step given CNNs' limited rotation invariance~\cite{goodfellow2009measuring}.

We train our network with supervised contrastive loss~\cite{kulis2013metric} on patches from the same and different keypoints. 
The CNN is composed of two convolution blocks with filter counts of 8 and 16.
Each block starts with two convolution layers with the same number of filters, a kernel size of three, and a Rectified Linear Unit (ReLU) activation function and ends with a max-pooling layer to downsample the width and height of the input.
After the convolutional layers, we flatten the output and use a dense layer with size 30 and an L2 normalization as our final output.
For training, we use the Multi-Similarity Loss~\cite{bursztein2021tensorflow} with the Adam optimizer and a learning rate of 0.0001.
We train the model for 15000 steps with a batch size of 4096 and early stopping to prevent overfitting.
The model weights with the smallest validation loss over the entire training run are used as the final weights of our model.
For the implementation, we use the Tensorflow Similarity library~\cite{MultiSimLoss2019}.

\begin{figure}[ht]
     \centering
    \vspace{-10px}
    \begin{minipage}[b]{0.45\textwidth}
        \centering
        \includegraphics[width=0.9\textwidth]{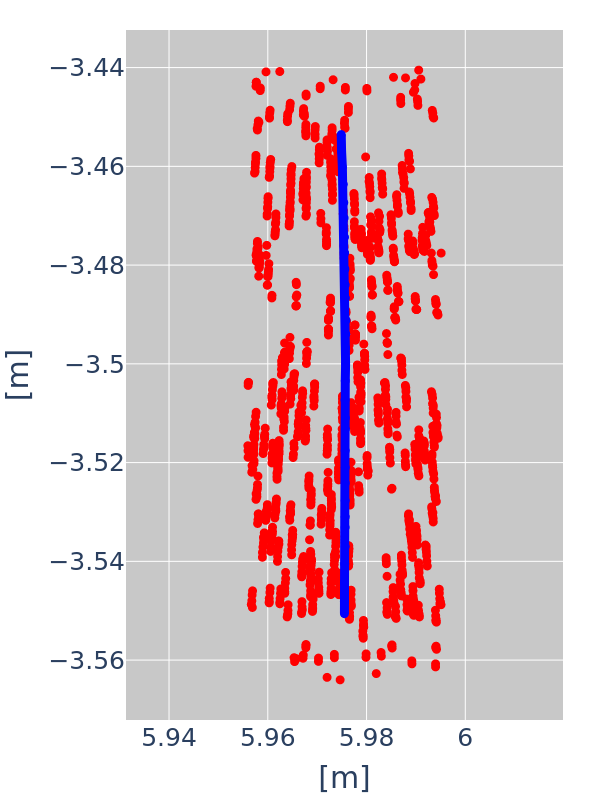}
    \end{minipage}
    \hfill
    \begin{minipage}[b]{0.45\textwidth}
        \centering
        \includegraphics[width=0.9\textwidth]{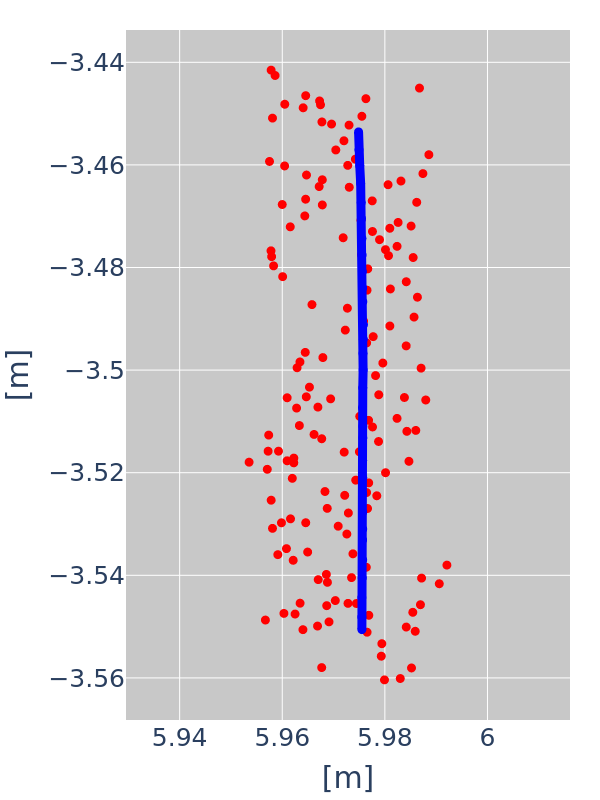}
    \end{minipage}
    \caption{Keypoints extracted from a run before clustering (left) and after (right).}
    \label{fig:pre_post_clustering}
    \vspace{-15px}
\end{figure}

The training dataset is generated by clustering keypoint patches (from another descriptor) that correspond to the same floor feature. 
Clusters with at least four members (as detailed in \autoref{sec:mapping}) yield four uniformly sampled images per cluster. 
An overview of the Segmentation Detector (SEG) and Similarity Descriptor (SIM) is shown in \autoref{fig:pipeline_overview}.

\subsection{Map Creation}
\label{sec:mapping}

We create a map database with keypoint descriptions and their position from the mapping runs.
The database is created with the following steps:

\begin{itemize}
    \item Image gathering with ground truth position.
    \item Outlier removal of ground truth positions.
    \item Process each image with a keypoint detection and description algorithm.
    \item Calculate the global position of each keypoint.
    \item Clustering and merging of keypoints.
    \item Storing and indexing of the descriptions in a database.
\end{itemize}

The floor images are captured with our robot as described in \autoref{dataset}.

Due to noise in the motion capture system, outliers are removed using a sliding window. 
The window moves over the ground truth positions along the recorded track.
In each window, we compute the mean value $\mu$ and the standard deviation $\sigma$ of the positions.
A new position $p_I$ is discarded if the deviation $d=|\mu - p_I|$ of this position from $\sigma$ exceeds the value $\alpha \cdot \sigma$ with a manually selected $\alpha$ of 0.8.

Clustering is required as each image is processed with a keypoint detector and descriptor to determine global keypoint positions from the image’s global coordinates and pixel locations. 
Given the zigzag capture pattern and overlapping images, the same color blob is often mapped multiple times. 

For clustering, we iterate through keypoints and select those within \SI{0.5}{\centi\metre} of each candidate, excluding keypoints from the same image or those already labeled. 
We then compute the cosine distance between descriptors and define a cluster as at least four keypoints with a cosine distance below 0.1. 
This approach is a simplified variant of DBSCAN \cite{ester1996density} that avoids transitive expansion for arbitrary shapes (\autoref{fig:pre_post_clustering}).

\subsection{Position Estimation}
\label{sec:position_estimation}
 
\begin{figure}[ht]
    \centering    
    \vspace{-15px}
    \smallskip
    \begin{minipage}[b]{0.49\textwidth}
    \centering
    \includegraphics[width=0.8\textwidth]{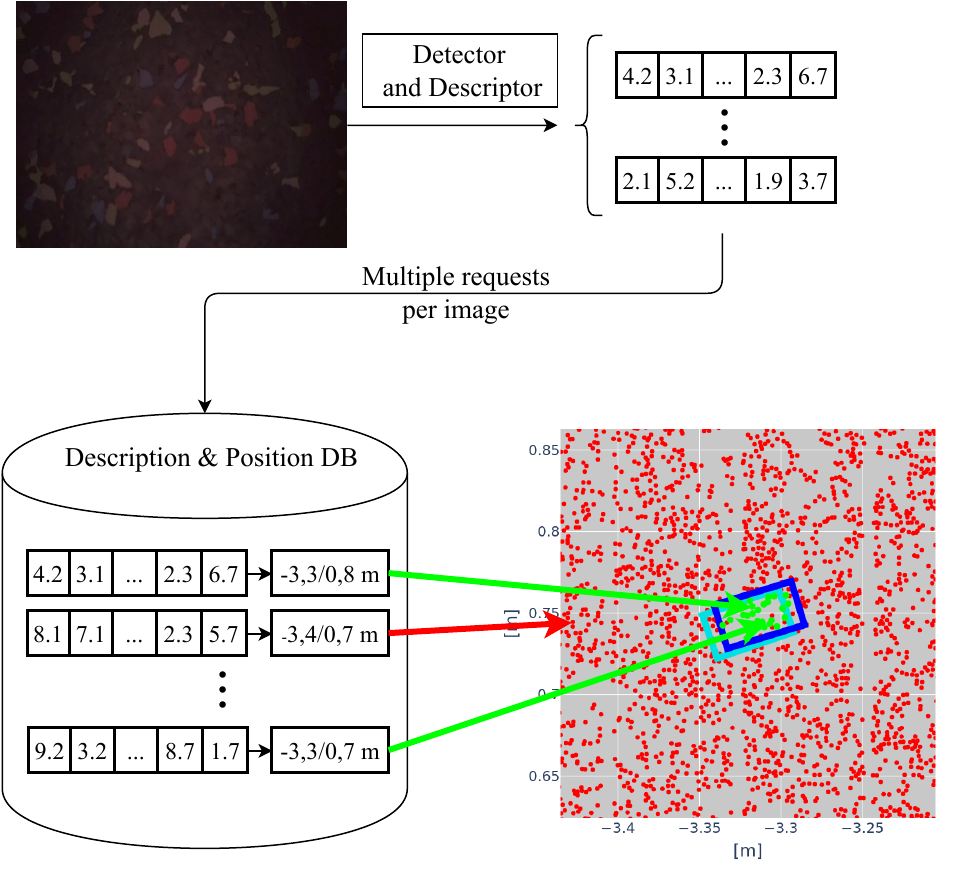}
    \caption{Pipeline for position estimation of a given image.}
    \label{fig:inf}
    \end{minipage}
    \hfill  
    \begin{minipage}[b]{0.49\textwidth}
    \centering
    \includegraphics[width=0.8\textwidth]{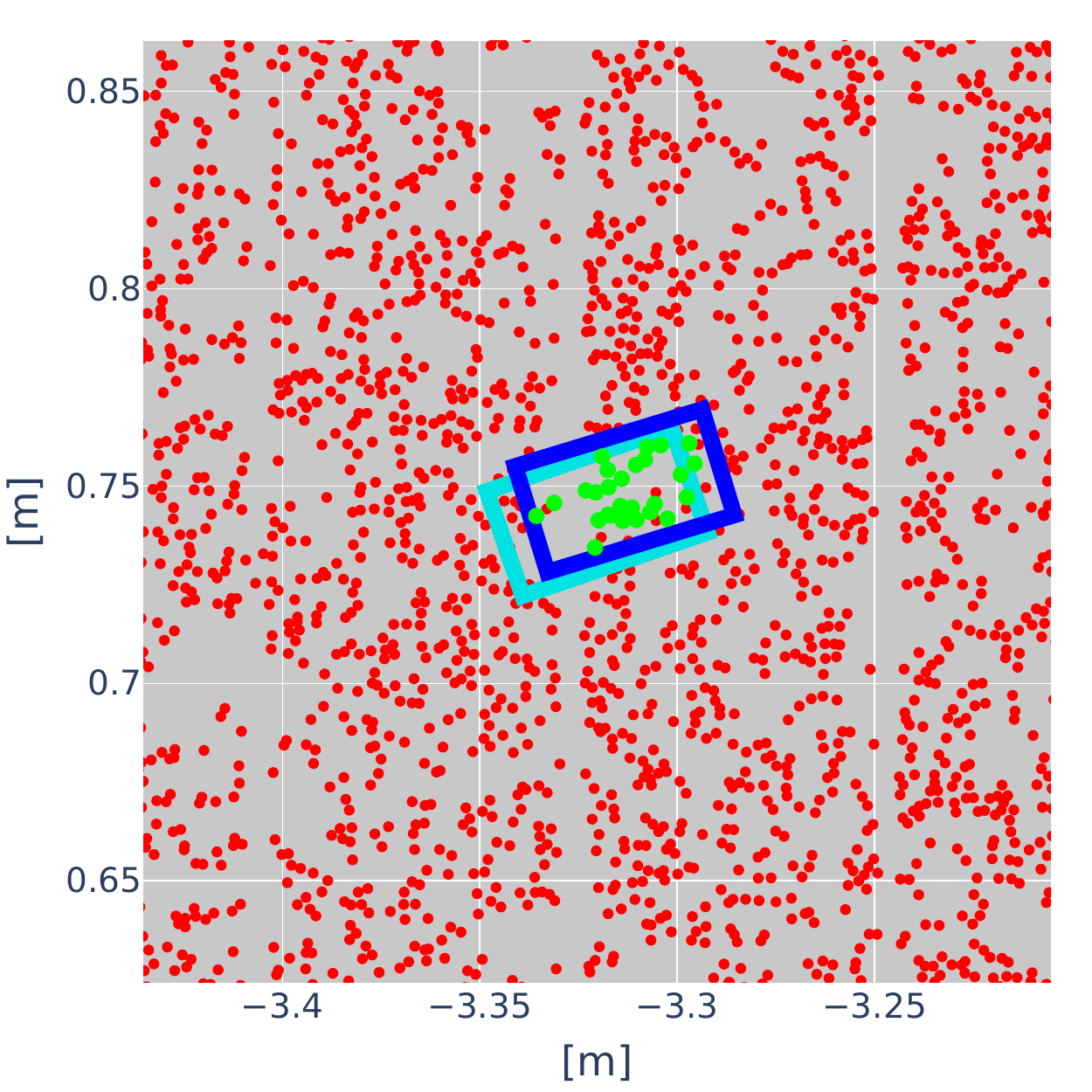}
    \caption{
    Database map: blue rectangle: estimated position, cyan: ground truth; red dots indicate stored keypoints, green dots show queried matches.
    }
    \label{fig:position_estimation}
    \vspace{-15px}
    \end{minipage}
\end{figure}

The position estimation predicts the position of an image independently without any prior information about previous positions and thus solves the kidnapping robot problem with every prediction.
An overview of the framework is shown in \autoref{fig:inf}.

For each image, keypoints and descriptors are obtained as in the map-creation process, and an approximate k-nearest neighbor search using TensorFlow Similarity~\cite{bursztein2021tensorflow} retrieves 20 comparable keypoints from the map database. 
Despite many spurious matches, the location with the highest match density (the mode) likely corresponds to the true position; hence, the keypoint with the most nearby matches is selected.

Subsequently, the optimal Euclidean transformation aligning the image keypoints with their map counterparts is computed using the filtered matches.
RANSAC is utilized to obtain the optimal transformation from the noisy data since some matches may still be mismatched~\cite{choi1997performance,fischler1981random}.
For RANSAC, we utilize a minimum sample size of three, a residual threshold of 0.002, and a maximum of 100 trials.
This transformation yields the predicted image position; if too few filtered matches exist or RANSAC fails, no prediction is made. \autoref{fig:position_estimation} illustrates a successful localization.

\section{Evaluation}
\label{sec:eval}

With the evaluation, we pursue the following goals: we evaluate the KOALA framework with various combinations of detectors and descriptors, as well as our Segmentation Detector and Similarity Descriptor, and we test our dataset of the feature-rich industrial floor.
We benchmark against baselines inspired by Kozak et al.~\cite{kozak2016ranger} and Zhang et al.~\cite{zhang2019high}.

The framework is evaluated using four evaluation runs for each of the three areas of the dataset with sizes of \SI{4}{\metre\squared}, \SI{36}{\metre\squared}, and \SI{144}{\metre\squared}.

We use multiple metrics to measure the quality of the KOALA framework.
For the prediction success rate (PSR), we consider the localization a success if the localization framework does not fail.
For the true success rate (TSR), a successful localization is defined as the predicted position being no more than \SI{10}{\centi\metre} and \SI{20}{degree} away from the ground truth position and rotation.
The mean distance and the mean rotational delta are calculated for each successful localization between the predicted and ground truth positions.
To obtain fewer false-positive predictions, it is also crucial for the PSR to be near the TSR.

We use SIFT and \mbox{CenSure} combinations as baseline detector/descriptor approaches, as they have shown strong performance in this field~\cite{kozak2016ranger,zhang2019high}. 
To evaluate KOALA, we compare these baseline approaches with our learned detector/descriptor as an alternative.

In addition, we tested SIFT as a detector in combination with our descriptor to see how well this combination works on our dataset.

For the evaluation of \mbox{CenSure} and SIFT, we primarily rely on the default parameters provided by the OpenCV implementation. 
However, to ensure consistency in the number of detected features corresponding to the colored blobs in the image, we adjust specific parameters. 
For \mbox{CenSure}, the \textit{responseThreshold} is set to 9, while for SIFT, \textit{nfeatures} is adjusted to 70, and \textit{contrastThreshold} is set to 0.03.

\begin{table}[htbp]
\vspace{-15px}
\caption{Success rate, mean position and angle errors of multiple detector and descriptor combinations for the \SI{144}{\metre\squared} area.}
\vspace{-8px}
\label{tab:main_eval}
\begingroup
\renewcommand{\arraystretch}{1.25} 
\begin{tabular}{lcccc}
\hline
Method & \thead{True \\ Success Rate} & \thead{Predicted \\ Success Rate} & \thead{Position \\ Error} & \thead{Angle \\ Error} \\ \hline
\mbox{SIFT-SIFT} & \SI{32.7}{\percent} & \SI{34.3}{\percent} & \SI{0.017}{\metre} & \SI{2.8}{\degree} \\
\mbox{\mbox{CenSure}-SIFT} & \SI{7.6}{\percent} & \SI{12.1}{\percent} & \SI{0.004}{\metre} & \SI{2.6}{\degree} \\
\mbox{SIFT-SIM-3} & \SI{63.0}{\percent} & \SI{69.4}{\percent} & \SI{0.019}{\metre} & \SI{2.8}{\degree} \\
\mbox{SEG-SIM-1} & \SI{64.3}{\percent} & \SI{67.4}{\percent} & \SI{0.019}{\metre} & \SI{2.4}{\degree} \\
\mbox{SEG-SIM-2} & \SI{73.4}{\percent} & \SI{76.4}{\percent} & \SI{0.019}{\metre} & \SI{2.4}{\degree} \\
\mbox{SEG-SIM-3} & \SI{75.7}{\percent} & \SI{78.0}{\percent} & \SI{0.020}{\metre} & \SI{2.4}{\degree} \\
\end{tabular}
\endgroup
\vspace{-18px}
\end{table}

When training the Similarity Descriptor, it is important to use a good descriptor to get a better training dataset using clustering.
Therefore, we train three Similarity Descriptors iteratively, with SIFT generating the training dataset for \mbox{SIM-1}, \mbox{SIM-1} generating the training dataset for \mbox{SIM-2}, and \mbox{SIM-2} generating the training dataset for \mbox{SIM-3}.

\begin{figure}[ht]
    \centering
    \vspace{-15px}
    \begin{minipage}[b]{0.45\textwidth}
        \smallskip
        \centering
        \includegraphics[width=\textwidth]{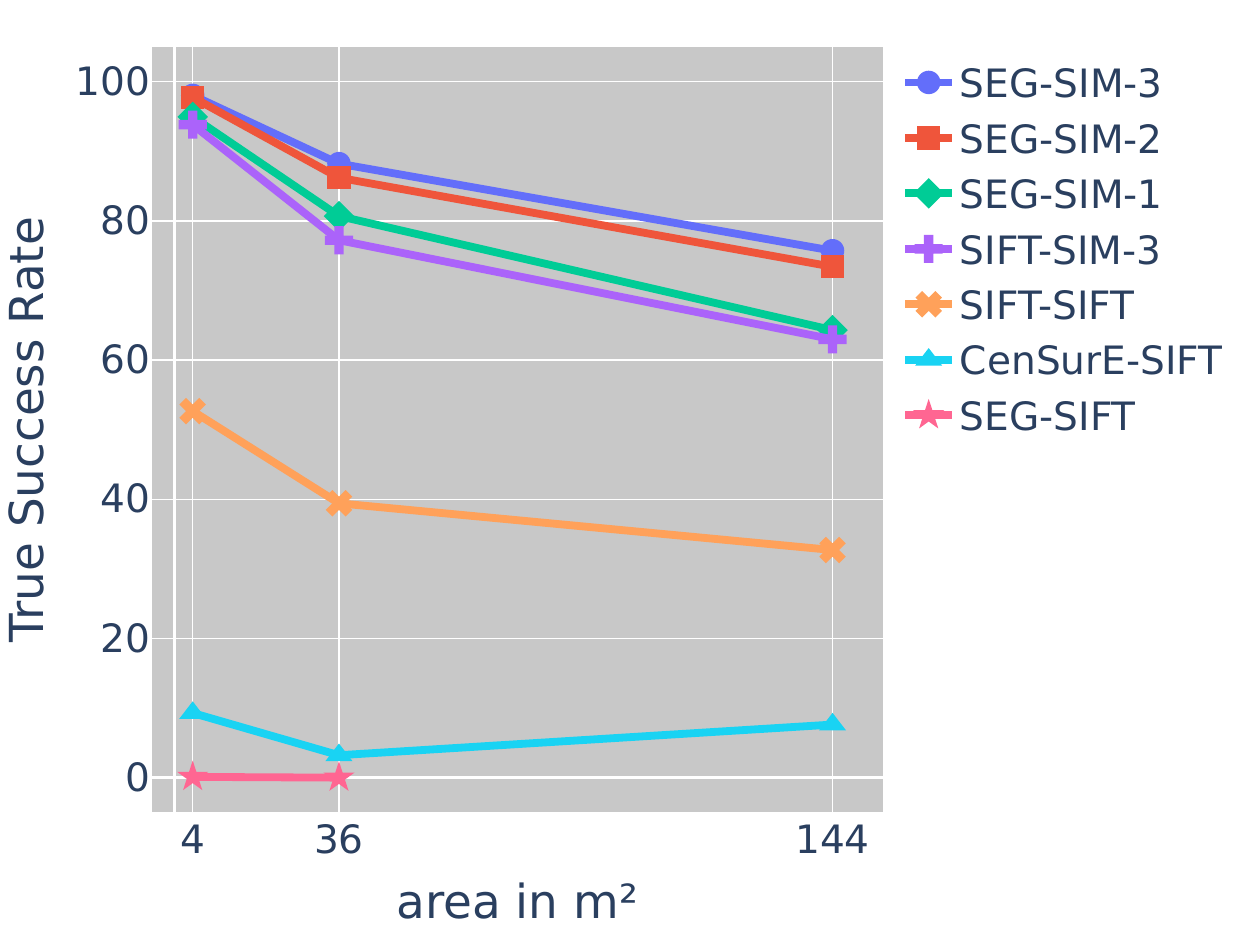}
        \caption{Evaluation of different detector and descriptor combinations for an area of \SI{4}{\metre\squared}, \SI{36}{\metre\squared} and \SI{144}{\metre\squared}.}
        \label{fig:plot_tsr_area}
    \end{minipage}
    \hfill
    \begin{minipage}[b]{0.45\textwidth}
        \centering
        \includegraphics[width=0.7\textwidth]{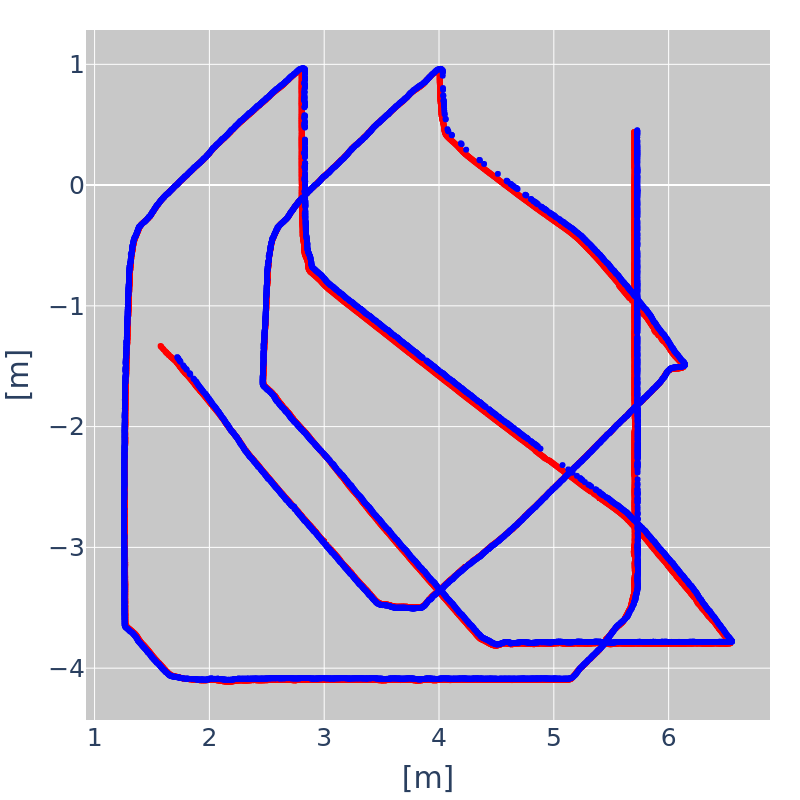}
        \caption{Trajectory of ground truth (red) and predicted (blue) robot positions on an evaluation run in the \SI{36}{\metre\squared} area using \mbox{SEG-SIM-3}.}
        \label{fig:pred_gt_trajectories}
        \vspace{-14px}
    \end{minipage}
    \vspace{-5px}
\end{figure}

The results of the detector and descriptor combinations on all three areas can be seen in \autoref{fig:plot_tsr_area} and for the biggest area in \autoref{tab:main_eval}.

According to \autoref{fig:plot_tsr_area}, the general performance of \mbox{\mbox{CenSure}-SIFT} and \mbox{SEG-SIFT} is unsatisfying.
This may be because many keypoints are in the middle of the color blobs, and SIFT does not seem to work well there since it only references a small area around each keypoint to generate its description.

It can also be seen in \autoref{tab:main_eval} that the true success rate performance increases with training multiple Similarity Descriptors iteratively from \SI{64.3}{\percent} to \SI{75.7}{\percent}.
Our learned approach also outperforms SIFT by a significant margin.
The results also show a good mean position and angle error for all tested methods.

\autoref{fig:pred_gt_trajectories} shows the ground truth and predicted position of all images in an evaluation run in the \SI{36}{\metre\squared} area using \mbox{SEG-SIM-3}.

Execution time is another essential metric for robot localization.
The timings were measured using an Nvidia Jetson Xavier NX board installed on the robot, with no further improvements to our framework, detector, or descriptor.
\mbox{SIFT-SIFT} takes \SI{0.59}{\second}, and \mbox{SEG-SIM-3} takes \SI{1.05}{\second} to estimate the position of one image.
Another distinction between the SIFT descriptor and our Similarity Descriptor is that our approach uses a smaller description dimension of~30, while SIFT uses 128.
This means that our descriptions are about four times smaller and therefore take up less space in the map database.

We conducted an analysis to evaluate the impact of specific components of our descriptor on position estimation performance. This investigation aims to assess the generalization capability of our descriptor while simultaneously addressing the timing overhead associated with the unified rotation algorithm.
Therefore we tested if the CNN encoder is able to learn rotation-invariant descriptions instead of using our unified rotation algorithm.
To achieve this, we have trained the CNN encoder with randomly rotating patches showing the same keypoints.

We also trained a CNN encoder to work with RGB instead of segmentation mask patches.
Our proposed change allows our descriptor to work with other detectors without generating a segmentation mask.
\vspace{-6px}
\subsubsection{Ablation Study}
The result of our ablation study is shown in \autoref{tab:ableation_eval}.
Our Similarity Descriptor can become rotation invariant at the expense of the true success rate.

The generalization to RGB images works quite well, performing almost as well as \mbox{SEG-SIM-3} with a learned rotation.
Both variants still significantly outperform \mbox{SIFT-SIFT}.

\begin{table}[htbp]
\vspace{-15px}
\caption{Evaluation of the \SI{144}{\metre\squared} run with \mbox{SEG-SIM-3} testing learned rotation invariance and RGB patches.}
\vspace{-8px}
\label{tab:ableation_eval}
\centering
\begingroup
\renewcommand{\arraystretch}{1.25}
\begin{tabular}{lcccc}
\hline

Method & Learned Rotation & RGB & True Success Rate\\ \hline
\mbox{SEG-SIM-3} & No & No & \SI{75.7}{\percent}\\
\mbox{SEG-SIM-3} & Yes & No & \SI{61.4}{\percent}\\
\mbox{SEG-SIM-3} & Yes & Yes & \SI{56.6}{\percent}\\

\end{tabular}
\endgroup
\vspace{-18px}
\end{table}
\section{Conclusion and Future Work}
 This paper shows the flexibility of our framework using different detector and descriptor combinations and that it can be used for ground localization of huge areas using the suitable detector and descriptor combination for the used floor.
We show that an industrial floor where colored granules were used for production instead of single-colored granules works well for scaled up ground localization.
It should be mentioned, that for localization, we use small images where each image only covers a fraction of $9.625 \cdot 10^{-6}$ of the \SI{144}{\metre\squared} area.

Even with such a small image, it is possible with the framework and our proposed detector and descriptor to localize with an accuracy of \SI{2}{\centi\metre} in 3 out of 4 cases.
The Segmentation Detector proposed by us, which is tailor-made for this floor, in combination with our Similarity Descriptor, outperforms other detector/descriptor combinations. 
Still, it has been shown that even these can achieve good results in variety with our framework.

To be able to use the framework productively in logistic halls, it makes sense to extend our system to a SLAM-like algorithm to eliminate the need for a motion tracking system to create a global map.

Furthermore, it is desirable to reduce the localization execution time to make it real-time usable and to make our detector usable for other floors as well.

%
%
%
\bibliographystyle{splncs04}
%
\bibliography{bibliography}
\end{document}